\documentclass{article}

\newcommand{\vbar}{\:|\:}

\newcommand{\fpq}{f_{pq}}
\newcommand{\wpq}{w_{pq}}

\newcommand{\tx}[2][1]{\ifthenelse{\equal{#1}{}}{f_{\hspace{-.04em}#2}}{f_{\hspace{-.04em}#2:\text{\fontsize{6}{6}\selectfont #1}}}}

\newcommand{\bx}{\mbox{\boldmath $x$}}
\newcommand{\by}{\mbox{\boldmath $y$}}
\newcommand{\bz}{\mbox{\boldmath $z$}}

\def\calL{{\cal L}}

\def\calP{{\cal P}}

\def\calS{{\cal S}}

\newcounter{transformation} 

\def\VVst{V_{st}}

\def\calP{{\cal P}}

\def\calS{{\cal S}}

\def\calL{{\cal L}}

\def\wpq{{w_{pq}}}

\usepackage{times}
\usepackage{epsfig}
\usepackage{graphicx}
\usepackage{amsmath}
\usepackage{amssymb}
\usepackage{mdwlist}

\usepackage{times}
\usepackage{epsfig}
\usepackage{graphicx}
\usepackage{amsmath}
\usepackage{amssymb}
\usepackage{algorithm}
\usepackage{algorithmic}
\usepackage[algo2e,titlenotnumbered,noline,ruled,longend]{algorithm2e}
\usepackage{slashbox}

\def\bx{\mathbf{x}}

\newcommand{\exclude}[1]{}

\begin{document}
\title{Efficient Graph Cut Optimization for Full CRFs with  Quantized Edges}

\author{Olga Veksler}

\maketitle

\begin{abstract}
   Fully connected pairwise Conditional  Random Fields (Full-CRF)  with Gaussian edge 
weights can achieve superior results compared to sparsely connected CRFs. However, traditional methods for Full-CRFs are too  expensive. Previous work develops efficient approximate optimization based on mean field inference, which is a local optimization method  and can be far from the optimum. We propose efficient  and effective optimization based on graph cuts for  Full-CRFs with  {\em quantized} edge weights.  To quantize edge weights, we partition the image into superpixels  and assume that the weight of an edge between any two pixels depends only on  the superpixels these pixels belong to. Our quantized edge CRF is an approximation to the Gaussian edge CRF, and gets closer to it as  superpixel size decreases.  Being an approximation, our model offers an intuition about the regularization properties of the Guassian edge Full-CRF.  For efficient inference, we first consider the two-label case and develop an approximate method based on transforming the original problem into a smaller domain. Then we handle  multi-label CRF by showing how to implement expansion moves. 
In both binary and multi-label cases, our solutions have significantly lower energy compared to that of mean field inference.  We also show the effectiveness of our approach  on semantic segmentation task.  \end{abstract}

The  work in~\cite{Koltun:DenseCRF} popularized  Fully Connected pairwise Conditional  Random Fields (Full-CRF). A Full-CRF   models long-range interactions by connecting every pair of pixels. It achieves superior results~\cite{Koltun:DenseCRF}  compared to sparsely connected CRFs.

Traditional discrete optimization methods that work well for sparsely connected
CRFs, such as graph cuts~\cite{BVZ:PAMI01} or TRWs~\cite{GTRWS:arXiv12},
are too  expensive for Full-CRF, as the number of potentials is quadratic
in the image size.  Taking advantage of the special properties of Gaussian edge weights, in~\cite{Koltun:DenseCRF} they develop an approximate optimization algorithm that is sublinear in the number of pairwise potentials.  It is based on mean field inference~\cite{Koller:2009:PGM} and approximate Gaussian filtering~\cite{Paris:IJCV:Filtering}.

It is well known that mean field inference, although efficient, is a local technique  and its solution can be arbitrarily far from the optimum.  For example,  in~\cite{Weiss01comparingthe} they compare  Belief Propagation (BP) to  mean field, and conclude  that mean field is inferior. BP is not the best performing  optimization method itself~\cite{kappes-2013} for loopy graphs.  Discrete optimization methods based on move-making with graph cuts work significantly better~\cite{kappes-2013}.

There are numerous extensions  to the original algorithm of~\cite{Koltun:DenseCRF}.
In~\cite{FullCRF:ICML2013} they extend their previous work to ensure convergence.
In~\cite{Chen:CVPR2012} they propose to augment  CRFs with object spatial relationships and develop optimization approach  based on quadratic programming relaxation.   
In ~\cite{IJCV:Vineet} they show how to incorporate higher order interaction terms.  
In~\cite{CRF:ECCV:2016} they propose continuous relaxation for optimization. The
approach in~\cite{DBLP:conf/eccv/BarronP16} speeds up the bilateral solver which further improves  the overall efficiency of the mean field  algorithm.
Full-CRFs  are gaining  more popularity because they can be combined with 
CNNs~\cite{Zheng:CNN_CRF,SchwingU15,ChenPKMY14, chen2015learning,jampani:cvpr:2016} in a unified framework.

The goal of our work is to develop a better optimization algorithm for a Full-CRF model.
We focus on the commonly used Potts model~\cite{BVZ:PAMI01} for pairwise potentials.
For  Potts model, the expansion algorithm~\cite{BVZ:PAMI01} is a popular choice for sparsely connected CRFs  due to its
 efficiency and quality trade-off~\cite{kappes-2013}. In fact the expansion algorithm has the best approximation factor for the case of Potts model, namely a factor of two.  This motivates us  to develop  expansion moves approach for  inference in 
Full-CRFs with Potts potentials. 
However, direct application of expansion is not feasible due to the  quadratic number of pairwise potentials.

Similar to~\cite{Koltun:DenseCRF} who restrict the form of allowed edge weights to be Gaussian, to obtain a Full-CRF model that can be optimized efficiently, we also restrict the edge weights to a certain form. 
In our model, we assume that image pixels have been tessellated into superpixels, and  the weight of an edge between two pixels depends only on the superpixels these pixels belong to. Our model is an approximation 
to  Gaussian edge Full-CRF~\cite{Koltun:DenseCRF},  and approaches it as superpixel size gets smaller, see Sec.~\ref{sec:GCRF}.  Being an approximation, our model offers novel insights into the regularization properties of the  Full-CRF in ~\cite{Koltun:DenseCRF}. 
We call our model  {\em quantized edge} Full CRF, since intuitively, it quantizes the Gaussian edge weights into bins.  
Quantized edge  assumption allows us to transform a large binary labeling problem into a much smaller multi-label problem that can be efficiently solved with graph cuts.

We first develop optimization for the case of two labels, i.e. binary Full-CRF, and then extend to the multi-label case with expansion moves.  
Inspired by~\cite{ Felzenszwalb:CVPR2010},  we  transform our problem into a reduced domain at the cost of introducing a larger number of labels. 
In particular, we reformulate the problem on the domain of superpixels~\cite{levinshtein2009turbopixels,Veksler:ECCV2010,achanta2012slic}. A naive approach would collapse all pixels in the same superpixel into a single entity, and 
then apply the standard expansion algorithm~\cite{BVZ:PAMI01}. However, this approach would only produce a coarse solution at the level of superpixels. Instead, we  change the label space from binary to  multilabel in order to encode different label assignments to   pixels  inside a superpixel.  Thus we produce  a solution in the original pixel space. 

Next we  extend our binary quantized edge Full-CRF optimization to the multi-label case by applying expansion moves.    We design a transformation that reduces an expansion move to the energy type 
required by our binary Full-CRF optimization. 

In addition to effective optimization, another advantage of our approach is that all edge costs are completely accounted for, no matter how small their weights are. This is unlike most other
methods for Full-CRF inference that disregard small weight edges. 

We evaluate our algorithm on  semantic image segmentation. 
We show that for the binary case, we  achieve the global minimum in the overwhelming majority of cases. 
For the multi-label case, our algorithm significantly outperforms mean field inference  especially as the
strength of the regularization is increased.

This paper is organized as follows. In Sec.~\ref{sec:prelim} we formulate our energy and explain its connection to the Gaussian edge model of~\cite{Koltun:DenseCRF} . 
In Sec.~\ref{sec:binary} we address optimization of binary Full-CRFs. 
In Sec.~\ref{sec:multi} we explain how to implement the expansion algorithm in the case of multi-label CRFs.
In Sec.~\ref{sec:efficientICM} we develop efficient mean field and ICM implementation for our quantized edge Full-CRF model. 
The experimental results are in Sec.~\ref{sec:experiments} and conclusion in Sec.~\ref{sec:conclusion}.

\section{Energy Function}
\label{sec:prelim}
In this section we formulate the energy function for our quantized edge Full-CRF model. 
Let $\calP$ be the set of image pixels, and $x_p \in \calL$ be the label  assigned to pixel $p$. 
Let  $\bx= (x_p\vbar p\in \calP)$ be the assignment of labels to all pixels.  
We wish to minimize
\begin{equation} 
f(\bx)=\sum_{p\in  \calP} f_p(x_p) + \sum_{p,q\in \calP} \fpq (x_p,x_q),
\label{eq:main-energy}
\end{equation}
where $$\fpq(x_p,x_q) =  \wpq \cdot  [x_p\neq x_q].$$ 
The unary terms $f_p(x_p)$ are the cost of assigning pixel $p$  to label $x_p$. They are usually known a-priori or learned from the data. The pairwise terms $\wpq \cdot [x_p\neq x_q]$ impose a penalty of $\wpq$ 
whenever pixels $p,q$ are not assigned to the same label.  The pairwise terms are used to regularize a labeling. In Full-CRFs, the summation
in Eq~\ref{eq:main-energy} is over all pairs of pixels in the image. Thus the number of pairwise terms is quadratic in image size.   

We assume an image is partitioned into superpixels. 
In~\cite{Koltun:DenseCRF}, $\wpq$ are based on Gaussian weighting of color
and spatial distance of pixels $p$ and $q$. Our edge weights are modeled similarly, 
but are based on superpixels. 
That is $\wpq$ is based on Gaussian weighting of color and  spatial distance of
 superpixels that contain pixels $p$ and $q$. This quantizes edge weights and  leads to large computational gains.

\begin{figure}
\begin{center}
 \includegraphics[width= 0.4\textwidth]{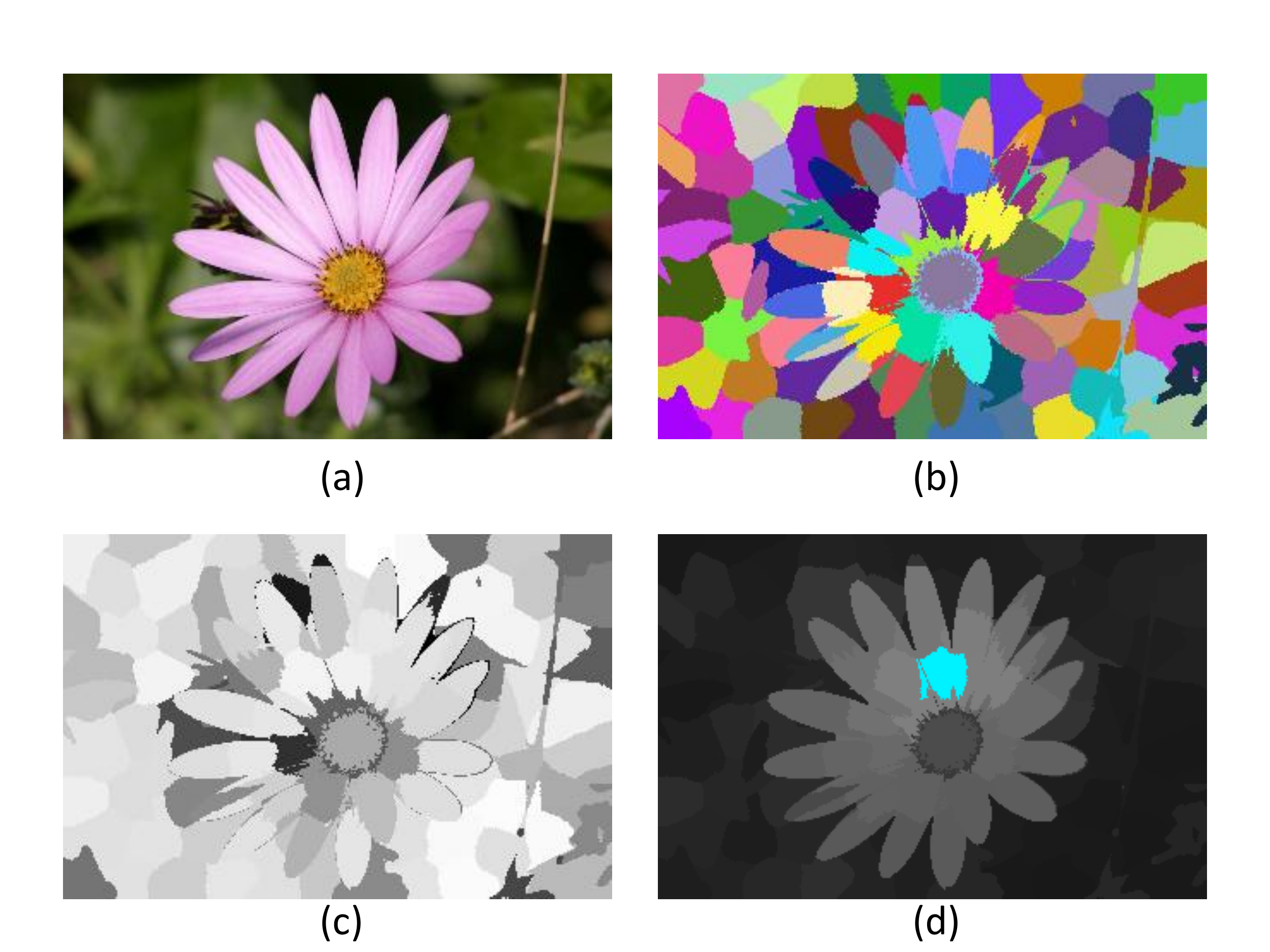}
\caption{Illustrates edge weights $\wpq$. Input image is in (a), superpixels computed  with~\cite{achanta2012slic} are in (b). In (c) we illustrate the weight strength between pixels inside the same superpixel. Brighter  intensities correspond to stronger edge weights.  In (d) we illustrate the strength of the edges that connect a pixel inside the  superpixel highlighted with blue and  the pixels inside other superpixels.  }
\label{fig:superpix}
\end{center}
\end{figure}

 Let $S(p)$ be the  integer index of the superpixel that pixel $p$ belongs to.  Let  $\mu_p$ be the intensity mean and $\sigma_p$ the intensity variance inside superpixel $S(p)$. Note that if $S(p)=S(q)$, then $\mu_p = \mu_q$ and $\sigma_p = \sigma_q$. 

We divide all edges into {\em internal} and {\em external}. Internal edges connect pixels that lie within the same superpixel. External edges connect pixels that lie in different superpixels. 
First we define edge weights for internal edges, i.e. the case when $S_p = S_q$:
\begin{equation}
\label{eq:wpqInside}
\wpq = \lambda_1 \cdot  exp \left (  -\frac{\sigma_p^2}{2  \beta_1^2 } \right ).
\end{equation}
In Eq.~\ref{eq:wpqInside}, we use intensity variance inside a superpixel for determining the edge strength. The higher is the variance, the smaller are the weights of edges inside that
superpixel. Intuitively, this corresponds to letting superpixels with higher variance to break across different labels more easily, since a higher variance superpixel is more likely to cross object boundaries.
Fig.~\ref{fig:superpix}(c) illustrates  internal edge weights. 
The higher variance superpixels are illustrated with darker intensities.  

Next we define the weights for external edges, i.e. the case when
  $S(p) \neq S(q)$:
\begin{equation}
\label{eq:wpqOutside}
\wpq = \lambda_1    \cdot exp \left (  -\frac{ ||d_p-d_q||^2}{2  \beta_2^2 } \right )+ 
 \lambda_2 \cdot  exp \left (  -\frac{ ||\mu_p-\mu_q||^2}{2  \beta_3^2 } \right ),
\end{equation}
where $d_p$ is the center of superpixel $S(p)$. The larger is the difference between the superpixel means, the smaller is  $\wpq$. The more distant two superpixels are, the smaller is  $\wpq$. 
Parameters $\lambda_1, \lambda_2,  \beta_1, \beta_2$ and $\beta_3$ are estimated from the training data. 

The external edge weights between one  superpixel (highlighted in blue) and all other superpixels are illustrated in~Fig.\ref{fig:superpix}(d).
Larger  edge weights are to the pixels that have similar color and are closer to the blue superpixel.

\subsection{Connection to Gaussian Edge Full CRF}
\label{sec:GCRF}
Our quantized edge model is an approximation to the Gaussian edge Full CRF~\cite{Koltun:DenseCRF}. As superpixels grow smaller (to one pixel in the limit)
and  $\beta$s in Eq.~\ref{eq:wpqOutside} larger, the edge weights defined by
Eq.~\ref{eq:wpqInside} and Eq.~\ref{eq:wpqOutside} approach the edge weights 
of ~\cite{Koltun:DenseCRF}. 

\begin{figure}
\begin{center}
 \includegraphics[trim={3cm 8cm 4cm 8cm},clip,width =0.4\textwidth]{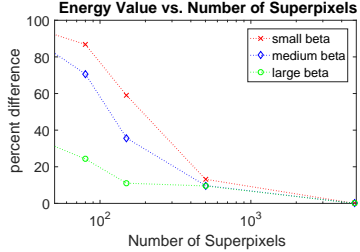}
\caption{Comparison of Gaussian edge model~\cite{Koltun:DenseCRF} with our
quantized edge model. On the $y$ axis we plot percent average relative difference of our energy from energy of the Gaussian 
edge CRF model. Increasing number of superpixels and increasing $\beta$ parameters
in Eq.~\ref{eq:wpqInside},\ref{eq:wpqOutside}
result in smaller relative percentage difference.  }
\label{fig:pixSuperP}
\end{center}
\end{figure}

We experimentally evaluate the convergence rate. 
We collected a set of 100 images of size $70$ by $70$ cropped from PASCAL 
dataset~\cite{Everingham10}, validation fold. 
We computed  pairwise energy of the ground truth labeling for our model and the one in~\cite{Koltun:DenseCRF} 
We omitted unary energy  terms since they are identical between the two models. 
When collecting image crops, we ensured that the ground truth labeling for the crop is not trivial, requiring the most frequent ground truth label to occupy less than two thirds of the image. 
We vary the number of superpixels and the width of the Gaussian parameters $\beta$ in  Eq.~(\ref{eq:wpqInside}) and~(\ref{eq:wpqOutside}).  We used $4900$, $500$, $150$, $80$, $50$ superpixels, where
energy with $4900$ superpixels is  equal to the Gaussian CRF energy in~\cite{Koltun:DenseCRF}. 
 We computed the relative difference (in percent) of the energy with $n$ superpixels from the energy with $4900$ superpixels, averaged over all images. 
The larger is  $\beta$, and the larger is the number of superpixels,
the closer is our model to the one in~\cite{Koltun:DenseCRF}, see Fig.~\ref{fig:pixSuperP}. 

The regularization properties of sparsely connected CRFs are well known, in particular, sparse CRFs offer boundary (length) regularization ~\cite{DBLP:conf/iccv/BoykovK03}.
In contrast, the regularization properties of Full-CRFs are not well understood. Most 
users of Full-CRFs make either  obvious statements, i.e. that Full-CRFs  model long-range interactions, or observational statements from the experiments, i.e. 
Full-CRFs preserve fine detail in the image. 

Being an approximation, our model offers an insight into the regularization properties of the model in~\cite{Koltun:DenseCRF}.
As we explain in Sec.~\ref{sec:binary}, assignment of pixels to a label inside each superpixel depends only on the unary terms
and the size (volume) of the split of pixels between the different labels, with a smaller split penalized less. 
This helps to explain why~\cite{Koltun:DenseCRF}  preserve fine detail. 
If a subset $T$ of pixels inside a superpixel has a strong unary preference for a label different from the rest of the pixels inside that superpixel, then the cost of  splitting $T$   from its superpixel, besides the unary terms,  depends only on the number of pixels in $T$. The shape of $T$ has 
no effect,  whether it is compact or  irregular, the cost is the same. 
 Thus fine structure can split off from the rest of the pixels inside a superpixel without a large penalty, provided its pixels have a strong unary preference for a different label. In contrast, with length based regularization, fine structure would have to pay a significant cost for its relatively long boundary.

\section{Optimizing Full CRFs: Binary Case}
\label{sec:binary}

We now explain our efficient optimization algorithm for the case when the energy in Eq.~(\ref{eq:main-energy}) is binary,
i.e. $\calL=\{0,1\}$. 
Without loss of generality,  we assume that $f_p(0)=0$ for all pixels $p\in\calP$, and $f_p(1)$ can be positive or negative. Any energy function  can be transformed to this form by subracting $f_p(0)$ from both 
$f_p(0)$ and $f_p(1)$ for all $p\in\calP$.  The new energy differs from the old one up to an additive constant.

Inspired by~\cite{ Felzenszwalb:CVPR2010}, we  transform our optimization problem to a different domain in order to 
greatly reduce the computational cost. In~\cite{ Felzenszwalb:CVPR2010} they develop an optimization
approach that can find a global minimum for a certain type of energy functions formulated on 2D images.  
The original optimization problem is transfered to a much smaller 1D domain at the cost of an enlarged label space. Similarly, we  reformulate our optimization problem in a reduced domain at the cost of introducing a larger number of labels.

\begin{figure*}
\begin{center}
 \includegraphics[width= 0.8\textwidth]{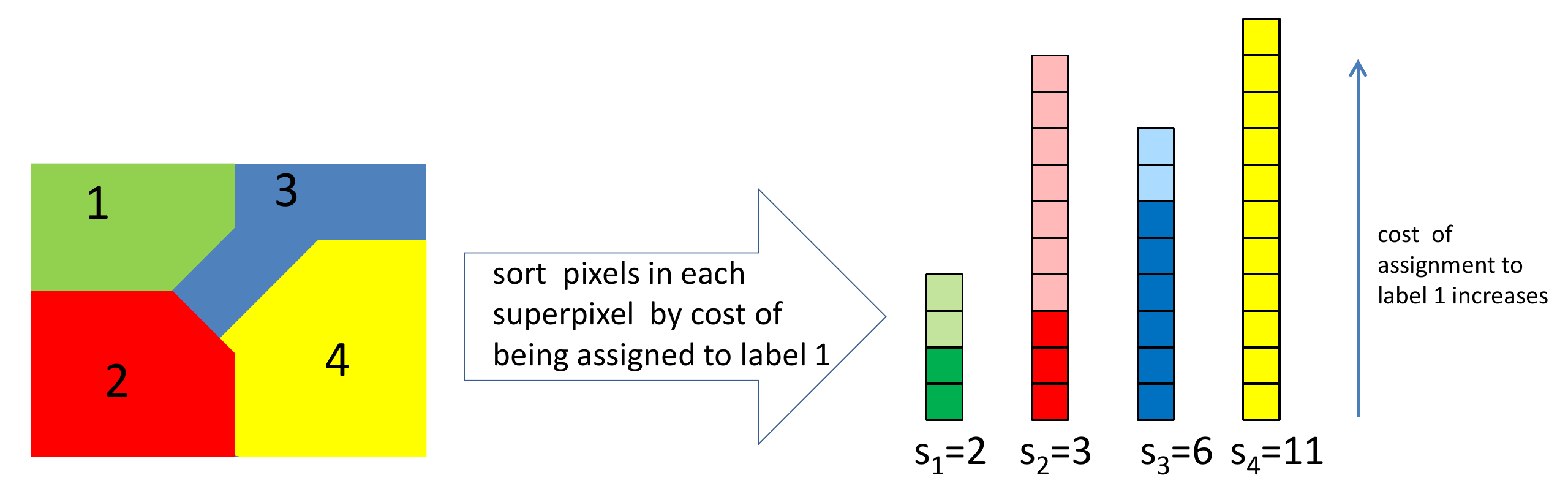}
\caption{
The transformation from the binary energy in the pixel domain to the multi-label energy in the superpixel domain.  The input image is partitioned into four superpixels, each containing 4, 10, 8 and 11 pixels, respectively. Superpixel $1$ can be assigned labels from the  set $\{0,1,...,4\}$, and similarly for the other three superpixels. 
We vertically stack the pixels in each superpixel in order of their preference to label $1$ in the original binary problem. Those that prefer label $1$ the 
most are on the bottom.   Superpixel $1$ is assigned to state $2$. This means that the $2$ of its pixels counting from the bottom are assigned to label $1$, and the rest to label $0$ in the original binary problem. Pixels assigned label $1$ in the original binary problem are shown with darker shade in each superpixel.  Similarly  for the other superpixels. All pixels in superpixel $4$ are assigned to label $1$, which corresponds to the largest label, namely label $11$ that superpixel $4$ can be assigned to. 
 }
\label{fig:transform}
\end{center}
\end{figure*}

We  formulate a new multi-label  optimization problem whose minimum corresponds to the
minimum of the two-label energy in Eq.~\ref{eq:main-energy}.
This process is illustrated in Fig.~\ref{fig:transform}.
 The domain for the new problem
is  the set of all superpixels $\calS$. Each superpixel $s\in\calS$ has its own set of labels $\calL_s = \{ 0, 1,..., n_s\}$ that can be assigned to it, 
where $n_s$ is the number of pixels in superpixel $s$.  Let  
 $y_s \in \calL_s$ be the label assigned to superpixel $s \in \calS$, and 
let $\by = (y_s, s \in \calS )$ be the assignment of labels to all superpixels. 

Let $P_s$ denote the set of all pixels that belong to superpixel $s$, and let $n_s$ be the number of pixels in $P_s$.
The correspondence between label $y_s$ and the binary labels of pixels in $P_s$ of the original  problem  in Eq.~\ref{eq:main-energy} is defined as follows. 
 If $y_s = k$, then exactly $k$  pixels in  $P_s$ are assigned to label $1$ and the rest are assigned to label $0$ in the original problem. 
The key observation that determines the correspondence is that these $k$ pixels must be those pixels in $P_s$ that have the smallest unary cost for label $1$. 

Indeed, consider superpixels $s,t$, and let $p,q \in P_s$ and $a \in P_t$.  
Let fixed $h$ be the number of pixels in  $P_t$ assigned to label $1$. 
Let us vary $k$, the number of pixels in $P_s$ that have label 1. 
The pairwise cost inside superpixel  $P_s$  is  $\wpq \cdot k  (n_s-k)$ 
and the pairwise cost between superpixels $s,t$ is  $w_{pa} \cdot\left (  k  (n_t-h) + (n_s-k)h\right ) $.
Thus the pairwise cost depends only on $k$, and so the optimal solution must assign to label
1 those $k$ pixels in $P_s$ that prefer label $1$ the most.

Fig.~\ref{fig:transform} illustrates the transformation of the binary energy in the 
pixel domain to the multi-label energy in the superpixel domain. 
For computational efficiency, we sort pixels in each superpixel in the increasing order of  unary cost of label $1$. This is done once in the beginning of the algorithm to avoid repeated sorting.

We now define the unary cost $g_s(y_s)$ of assigning label $y_s$ to superpixel $s$. 
 Let $o(p)$ be the sorted order of pixel $p$ in the superpixel  it belongs to. That is if $p$ has the smallest cost of being assigned to label $1$, then  $o(p)=1$. 
Then
\begin{equation}
\label{eq:g-unary}
g_s(y_s) =   {\wpq \cdot y_s  (n_s-y_s)} + \sum_{\substack{p\in P_s \\ o(p) \leq y_s} }f_p(1) ,
\end{equation}
where $p,q$ are any distinct pixels in $P_s$. 
The first term in Eq.~\ref{eq:g-unary} accounts for the pairwise terms of the original energy that depend only  on pixels inside superpixel $s$.  
The second  terms in Eq.~\ref{eq:g-unary} accounts for the unary terms of the original energy in
Eq.~\ref{eq:main-energy} that depend only  on pixels inside superpixel $s$.
Note that since $f_p(0)=0$ for any pixel $p$, it does not have to be accounted for.

Pairwise cost  for assigning labels $y_s$, $y_t$ to superpixels $s,t$ is 
\begin{equation} 
\VVst(y_s, y_t) =  \wpq \cdot \left ( y_s(n_t-y_t) + y_t(n_s-y_s)  \right ), 
\label{eq:vpq-new}
\end{equation}
where $p$ is any pixel in $P_s$ and $q$ is any pixel in $P_t$. This cost adds up how many
nonzero pairwise  terms of the original energy are  there between pixels in 
   $P_s$ and $P_t$.  These are the costs between  $y_s$ pixels of $P_s$  that are labeled as $1$ and
$n_t-y_t$ pixels of $P_t$ that are labeled as $0$, plus  the costs between  $n_s-y_s$ pixels of $P_s$  that are labeled as $0$ and
 $y_t$ pixels of $P_t$ that are labeled as $1$.

The complete energy is
\begin{equation} 
g(\by)=\sum_{s\in  \calS}g_s(y_s) + \sum_{s,t\in \calS} \VVst (y_s, y_t).
\label{eq:main-new}
\end{equation}

It is convenient to rewrite Eq.~\ref{eq:vpq-new} as 
\begin{eqnarray}
\VVst(y_s,y_t) &= & \wpq \cdot (y_s-y_t)^2  \nonumber  \\
                     &- & \wpq \cdot \left ( y_s^2 +y_s n_t - y_t^2 + y_t n_s   \right ).
\label{eq:vpq-new2}
\end{eqnarray}

We can add  $\wpq  (y_s n_t - y_s^2 )$ to the unary term of superpixel $s$, 
and $\wpq ( y_t n_s- y_t^2)$
to  the unary term of superpixel $q$. This leaves pairwise term 
$ V_{st}(y_s,y_t)=\wpq (y_s-y_t)^2$. Optimization of energies with quadratic pairwise terms can be solved exactly~\cite{Ishikawa:2003,Schlesinger:Report}.
Thus the energy in Eq.~\ref{eq:main-new} can be optimized exactly with the algorithm in~\cite{Ishikawa:2003}. However, this approach is only somewhat more efficient compared to
optimizing the original binary energy directly. This is due to the cost of constructing a graph with $n_t 
\cdot n_s$ edges for each pair of superpixels $s,t$. The total number of edges would be smaller by a factor that is roughly equal to the average superpixel size. This is still computationally expensive for a fully connected graph. 
Note that the algorithm in~\cite{AjanthanHS17} can be used for a memory efficient implementation of energy in  Eq.~\ref{eq:main-new}, however, computational efficiency is still too high for a Full-CRF.

Instead of optimizing Eq.~\ref{eq:main-new} exactly with the exact by  expensive construction~\cite{Ishikawa:2003}, we use the expansion algorithm~\cite{BVZ:PAMI01}. 
Expansion is an iterative optimization method that starts with some initial solution $\by$ and tries to improve
it by finding the optimal subset of superpixels to switch to some fixed label $\alpha$. The graph constructed
during expansion is only linear in the number of superpixels, which is very efficient. Several
iterations over all labels in $\calL^* = \bigcup_s {\calL_s}$ may be required. We found that the energy converges after 
one or two iterations in most cases. The small number of iterations required for convergence is probably due to the energy being relatively easy to optimize. 

When expanding on label $\alpha \in \calL ^*$, this label is infeasible for any superpixel $s$ with size less than $\alpha$.
For such superpixels, we set the unary cost for  $\alpha$ to infinity. 
We also found it helpful to perform expansion ``in reverse''. The intuitive meaning of expanding on $\alpha$
is that we are trying to assign the same number   $\alpha$ of pixels in every superpixel to label $1$ of the original binary problem. In this case, the penalty is small. Due to symmetry, it makes sense to expand on labels in reverse, i.e. switch the meaning of labels $0$ and $1$ of the original binary problem.
When expanding on label $\alpha$ in  ``reverse'', we are trying to assign the same number of pixels $\alpha$ to  the label $0$ in every superpixel. 
  Additional optimization  significantly improves the  quality.

Since $V_{st}$ is not a metric~\cite{BVZ:PAMI01}, expansion is not guaranteed to find the optimal 
subset of superpixels to switch to label $\alpha$ in our case.  However, in our experiments we 
almost always find the optimal solution,  see Sec.~\ref{sec:experiments}. 
We use the ``truncation trick'' from ~\cite{digital-tapestry} to handle non-submodularity of expansion

Note that if the unary terms were also convex, then the energy in Eq.~\ref{eq:main-new} could be optimized
with the jump moves proposed in~\cite{Veksler:phd,KolmogorovS09} without the need to construct a large graph. 
Our unary terms are not convex since we add $- y_s^2 +y_s n_t$ to them. Still, we 
 evaluated the jump moves and found them inferior to the expansion moves, see Sec.~\ref{sec:experiments}.

\section{Optimizing Full CRFs: Multi-label Case}
\label{sec:multi}
In Sec.~\ref{sec:binary} we explained our efficient algorithm for optimizing a quantized edge Full-CRF in case when the energy 
in Eq.~\ref{eq:main-energy} is binary. We now turn to   general multi-label case, i.e. $|\calL| \geq 2$. 

We use expansion algorithm for optimization, iterating expansions on labels in $|\calL|$. 
Each $\alpha$-expansion is implemented via optimization of a binary energy. Assigning pixel $p$ to label $0$ means that 
it stays with its old label $x_p$, while assigning label $1$ to pixel $p$  means that it switches its label to $\alpha$. 
Thus finding the best $\alpha$-expansion move on Full-CRF can be formulated as optimization of
a binary expansion  energy on Full-CRF. However, straightforward formulation of the binary expansion
energy results in an energy different from the form required in Sec.~\ref{sec:binary}. In this section we develop
a formulation that is in the form required in Sec.~\ref{sec:binary}

We start by describing the binary expansion energy.
 Let $\bx$ be the current labeling for which
we wish to find the optimal $\alpha$-expansion move. We introduce a binary
variable $z_p$ for each pixel $p$, and collect all these variables into vector $\bz = (z_p,p \in\calP)$.
The meaning of binary variables $z_p$ is as follows. If  $z_p = 0$, pixel $p$ stays with its old label.
If $z_p=1$, pixel $p$ switches its label to $\alpha$.

The unary terms $h_p(z_p)$ are as follows: $h_p(0) = f_p(x_p)$ and $h_p(1) = f_p(\alpha)$.
If pixel $p$ has label $\alpha$ in the current solution, then the new and the old labels for $p$ are the same.  In this case we prohibit assigning label $1$ to $z_p$ by setting $h_p(1) = \infty$, to ensure that the algorithm is correct. 

The pairwise terms for pixels $p$ and $q$ are 
\begin{equation}
h_{pq}(z_p,z_q)  =  \left \{  \begin{array}{cl}
				 0  & \mbox{if }  z_p =1  \mbox{, }   z_q =1 \\                         
				 0  & \mbox{if }  z_p =0  \mbox{, }   z_q =0 \mbox{ and } x_p = x_q \\                         
                           \wpq & \mbox {otherwise.} \\                                  
                  \end{array} \right .
\end{equation}

Before we can apply the method developed in  Sec.~\ref{sec:binary}, we need to make modifications to the energy $h(\bz)$, as it is
not of the form assumed in  Eq.~\ref{eq:main-energy}.
The problem is that in  Eq.~\ref{eq:main-energy}, the meaning of label $0$ is always the same, and if two pixels are assigned to label $0$, 
there is no pairwise cost.  In the case of
expansion, there may or may not be a pairwise cost, depending on whether these pixels have the same current label or not.

\begin{figure}
\begin{center}
 \includegraphics[width= 0.4\textwidth]{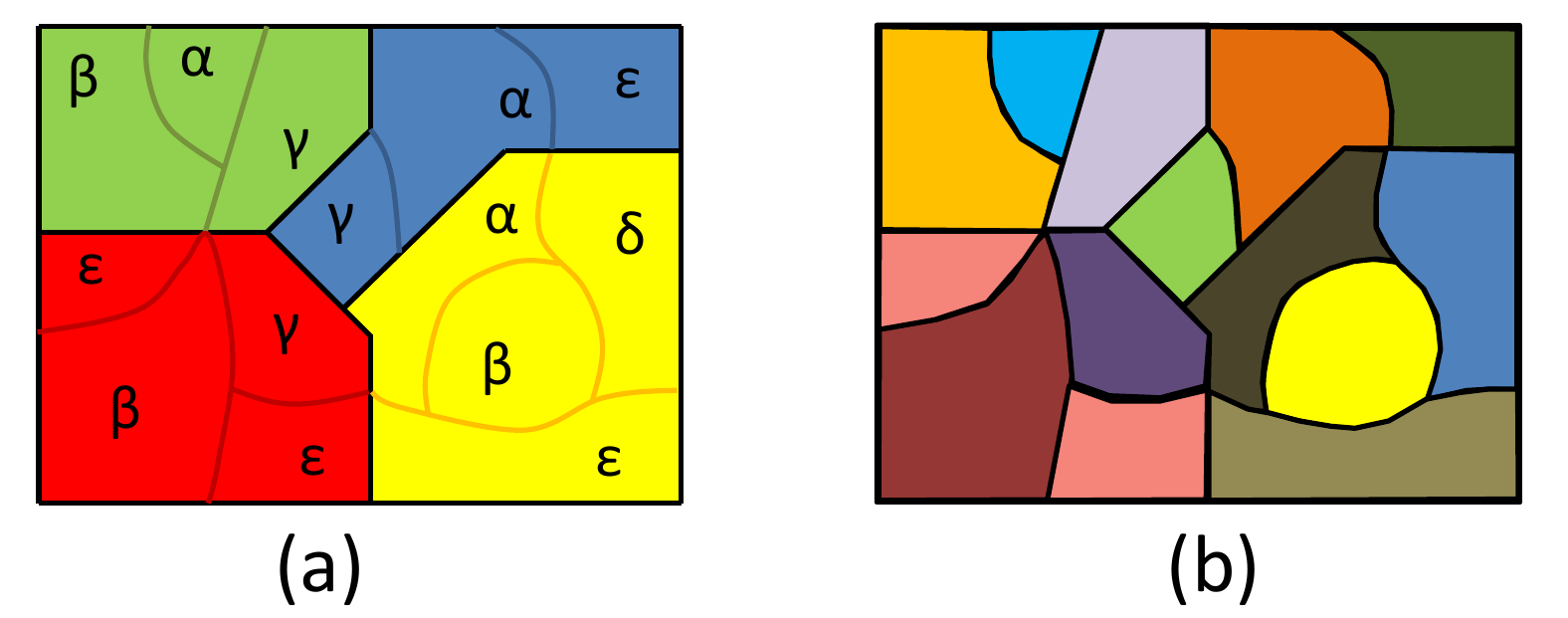}
\caption{
Illustrates formation of new superpixels: (a) for original superpixels shown with different colors. The pixels inside these superpixels have different labels, shown with greek letters; (b) shows the new superpixels formed by breaking the original four superpixels in (a) according to the current labeling. 
 }
\label{fig:conversion_lena}
\end{center}
\end{figure}

To convert the binary expansion energy $h(\bz)$ to the required form, first  we need
our superpixels to satisfy the  following property.   For any superpixel $s$ and any pixels $p,q \in P_s$, we need $x_p = x_q$.
Thus we split  the original superpixels we started with  further, according to labeling $\bx$, illustrated in
Fig.~\ref{fig:conversion_lena}.  If superpixel $P_s$ contains pixels currently labeled as $\alpha, \beta, \gamma$, 
it is split into three new superpixels each containing pixels that have the same current label. 
For simplicity, we will use the same old notation for the new superpixels.  So from now on, we assume that
superpixels are split and the new superpixels $P_s$ have the required property.

We formulate a new energy $d(\bz)$ equivalent to $h(\bz)$ as follows.  
The pairwise terms are 
$$d_{pq}(z_p,z_q) =v_{pq} \cdot [z_p \neq z_q],$$ where
\begin{equation}
v_{pq}  =  \left \{  \begin{array}{cl}
				    \wpq   & \mbox{if }  x_p  = x_q  \\                         
                           \frac{1}{2} \wpq   & \mbox {if } x_p  \neq x_q \\                                  
                  \end{array} \right .
\end{equation}
Thus in the new energy $d(\bz)$, the pairwise terms are of the form as  needed
in Eq.~\ref{eq:main-energy}.  The  fraction of 
$\frac{1}{2}$ ``underpayment'' in the case when pixels $p$ and $q$
do not have the same current labels is corrected by the unary terms $d_p$ and $d_q$, defined next.

The unary terms $d_p(z_p)$ are defined as follows
\begin{equation}
d_{p}(z_p)  =  \left \{  \begin{array}{ll}
                            h_p(0) +  \sum_{q   \in P \setminus \{p\} }  \frac{ \wpq }{2}  & \mbox {if }z_p = 0 \\                                  
				 h_p(1)  & \mbox{if }  z_p  =1. \\                         
                  \end{array} \right .
\end{equation}
This definition ensures that whenever $z_p = 0$, the cost involving the current label of pixel
$p$ is modeled correctly. It is straightforward to check that $d(\bz) = h(\bz)$ for all binary
vectors $\bz$. 

Thus our overall algorithm consists of two nested invocation of the expansion algorithm.
In the outer invocation, we iterate over the multilabel set $\calL$, calling expansion algorithm
for each $\alpha \in \calL$. In the inner invocation, we transform the binary expansion energy 
from the pixel domain to  the superpixel domain, and run the expansion algorithm over the new label set in the superpixel domain.

\section{Efficient ICM and Mean Field}
\label{sec:efficientICM}
We now explain how to implement ICM~\cite{Besag:JRSSB86}
and mean field~\cite{Koltun:DenseCRF}  inference for our quantized edge Full-CRF efficiently.
Unlike the inference approaches based on approximate filtering~\cite{Paris:IJCV:Filtering},
for our energy model, all the mean field iteration steps are exact.

\subsection{Efficient Implementation of  ICM}
There are two versions of ICM that we implement: pixel and superpixel level. Starting with an initial labeling,  the pixel level ICM iteratively switches the label of each   pixel to the one that gives the best energy decrease. This is repeated until convergence. The superpixel ICM is similar, except the labels of all pixels in a superpixel must switch to the same label.

Let us first consider pixel level ICM. 
To efficiently compute the best label, instead of computing the full energy, we only compute the decrease in the energy if a pixel  is switched to a new label. To compute the energy decrease efficiently, for each superpixel, we store how many pixels it has for each possible label. Given superpixel $s$, let $n^l_s$ be the number of pixels that have label $l$ in superpixel $s$ in the current labeling. 
Let $n_s$ denote the size of superpixel $s$. Then the label corresponding to the  best energy decrease can be computed in $O(mk)$ time for each pixel $p$, where $m$ is the number of superpixels and $k$ is the number of labels in $\calL$. This is because for each superpixel $s$, the weight between pixel $p$ and any pixel in $s$ is constant. 
Thus we can aggregate information over the blocks of pixels in each $s$ that have the same label.
In particular, let  the current label of pixel $p$ be $l$, and suppose we are considering switching pixel $p$ to
label $a$. Let  $w^s_p$ be the weight
between pixel $p$ and any pixel in superpixel $s$. If $s$ is the superpixel that $p$ belongs to,
then $w^s_p $ is the weight between pixel $p$ and any other pixel $q\neq p$,  $q\in s$.  The energy change is computed as
\begin{eqnarray}
\delta(p,l,a) &=& f_p(a)-f_p(l)  +\sum_{s\in\calS \setminus \{ S(p) \} }{  w^s_p \cdot (n^{l}_s - n^{a}_s) }  \nonumber \\
&+&  w^{S(p)}_p \cdot (n^{l}_{S(p)}-1 - n^{a}_{S(p)}),  \nonumber \\
\end{eqnarray}
where $\calS$ is the set of all superpixels, and $S(p)$ is the index of superpixel that contains $p$.

Computing $\delta(p,l,a)$ for one pixel and all labels $a\in\calL$ is $O(mk)$, where $m$ is the number of superpixels and $k$ is the number of labels. One iteration over all pixels is  $O(nmk)$, where $n$ is the number of pixels in the image.  Since the number of superpixels is much smaller than the number of pixels, this complexity  is much better than a naive $O(kn^2)$  implementation.

For superpixel ICM, we need to compute the cost of switching all pixels in superpixel $s$ from label $l$ to label $a$,  computed as
\begin{eqnarray}
\delta(s,l,a) &=& \sum_{p \in \calP_s  } (f_p(a)-f_p(l) )  \nonumber \\
&+&\sum_{t\in\calS \setminus \{ s \} }{  w^{st} \cdot (n_s  (1- \delta^a_t)  n_t- n_s    (1-\delta^l_t ) n_t), }  \nonumber 
\end{eqnarray}
where $\calP_s$ be the set of pixels in superpixel $s$,  $w^{st}$ is the cost of any edge between
a pixel in superpixel $s$ and a pixel in superpixel $t$. Also  $\delta^l_s=1$ if superpixel $s$ is currently assigned to label $l$, and 
$\delta^l_s=0$  otherwise. 

For one superpixel, $\delta(s,l,a)$ is computed in $O(k(n_s+m))$ time for all labels. One iteration 
consists of computing $\delta(s,l,a)$  for all superpixels. Thus complexity of one iteration is 
 $O(k(m+n))$, which is significantly better than the naive implementation.

\subsection{Efficient Implementation of  Mean Field Inference}

\begin{algorithm2e}
\caption{{\tt Mean Field Inference
\label{alg-meanfield}}}

\tcp{Initialization}

$Q_p(x_p) = \frac{exp{\left (-f_p(x_p) \right )}}{\sum_{l\in\calL}exp{\left (-f_p(l) \right )}}, \;\;\;\;\; \forall p\in \calP, \forall x_p \in \calL$

\Repeat{ until convergence    }{
\For{$p \in \calP, l \in \calL$}
{

\tcp{Message Passing}
$\tilde{Q}_p(l) = \sum_{q \neq p } {\wpq \cdot Q_q(l)  }$  
	
\tcp{Compatibility Transform}
$\hat{Q}_p(l) = \sum_{l' \in \calL } {[l' \neq l]  \cdot \tilde{Q}_p(l') }$

\tcp{Local Update}
$Q_p(l) =  exp {  \left (   -f_p(l) - \hat{Q}_p(l) \right )}$

\tcp{Normalization}
$Q_p(l) = \frac{Q_p(l) } {\sum_{l'\in \calL} {Q_p(l') } } $ 
}
}
\end{algorithm2e}

The mean field inference is summarized in Alg.~\ref{alg-meanfield}.
It consists of the initialization step and the iterations inside the {\bf{repeat}} loop. 
The only step which is costly if not carefully
implemented is the message passing stage, the first stage of the {\bf {repeat}} loop. 
To implement it efficiently, observe that for any pixels $p,r$ that are inside the same superpixel,
most of the summation terms when computing $\tilde{Q}_p(l)$ and $\tilde{Q}_r(l)$ are equal. Thus calculations of $\tilde{Q}_p(l)$ for pixels inside the same superpixel can be shared. 
 In particular, for each superpixel $s$ and label $l$ 
we first precompute
\begin{equation}
sum_e(s,l)  =  \sum_{t \in \calS \setminus \{ s\} } {w^{st} \sum_{p \in \calP_t } {  Q_p(l)  }}. 
\label{eq:sume}
\end{equation}
In Eq.~\ref{eq:sume}, $sum_e(s,l)$ is the cost of $\wpq \cdot Q_q(l)$ terms of the message passing stage
that go between pixel $p$ in superpixel $s$ and pixels $q$ that are outside of superpixel $s$. 
Thus $sum_e(s,l)$ is shared by all pixels inside superpixel $s$. 

Next we compute the internal sum 
\begin{equation}
sum_i(s,l)  =  \sum_{ p \in \calP_s } {w^{ss} \cdot {  Q_p(l)  }}. 
\label{eq:sumi}
\end{equation}
For any $p$ in superpixel $s$, $sum_i(s,l)$ in Eq.~\ref{eq:sumi} is almost what we need
to add to $sum_e(s,l)$ in order to get the correct expression for $\tilde{Q}_p(l)$.
The only problem is that it has one extra term, namely   $Q_p(l)$.
Therefore, to get the correct calculation,  for any $p$ in superpixel $s$, we compute 
\begin{equation}
\tilde{Q}_p(l)  = sum_e(s,l)+sum_i(s,l)  - w^{ss}  \cdot Q_p(l).
\end{equation}
Performing the calculation in Eq.~\ref{eq:sume} is $O(nk)$ for one superpixel and all labels. 
Performing it for all superpixels is $O(mnk)$. Calculating the 
sum in  Eq.~\ref{eq:sumi} is $O(nk)$ for all labels and superpixels. 
thus the total time to perform one iteration of message passing is $O(nkm)$. 
The compatibility transform stage is  $O(n k^2)$, and the other stages are less expensive. 
Thus the total cost of one iteration of the mean field inference is $O(n k^2 + nkm)$,
which is much less expensive than naive $O(n^2)$ implementation if the number of
superpixels is much less than the number of pixels.

\section{Experimental Results}
\label{sec:experiments}
 The main goal of our  experiments is to demonstrate that our approach has a superior optimization performance to that of the mean field 
inference~\cite{Koltun:DenseCRF} commonly used for optimization of Full CRFs. We also compare against ICM~\cite{Besag:JRSSB86}. 
Both ICM and mean field are implemented efficiently  
as  in Sec.~\ref{sec:efficientICM}.

We use PASCAL VOC2012 segmentation dataset~\cite{Everingham10}. The images are of size approximatedly $500\times 300$ and 
there are 21 object class labels.
For unary terms, we used a pre-trained CNN classifier from~\cite{Zheng:CNN_CRF},  available for downloading from~\cite{vlfeet}.
We use~\cite{achanta2012slic} to compute superpixels, approximately 200 per image. 

\subsection{Binary Full CRFs}

\begin{figure}
\begin{center}
 \includegraphics[trim={3cm 8cm 4cm 8cm},clip,width =0.23\textwidth]{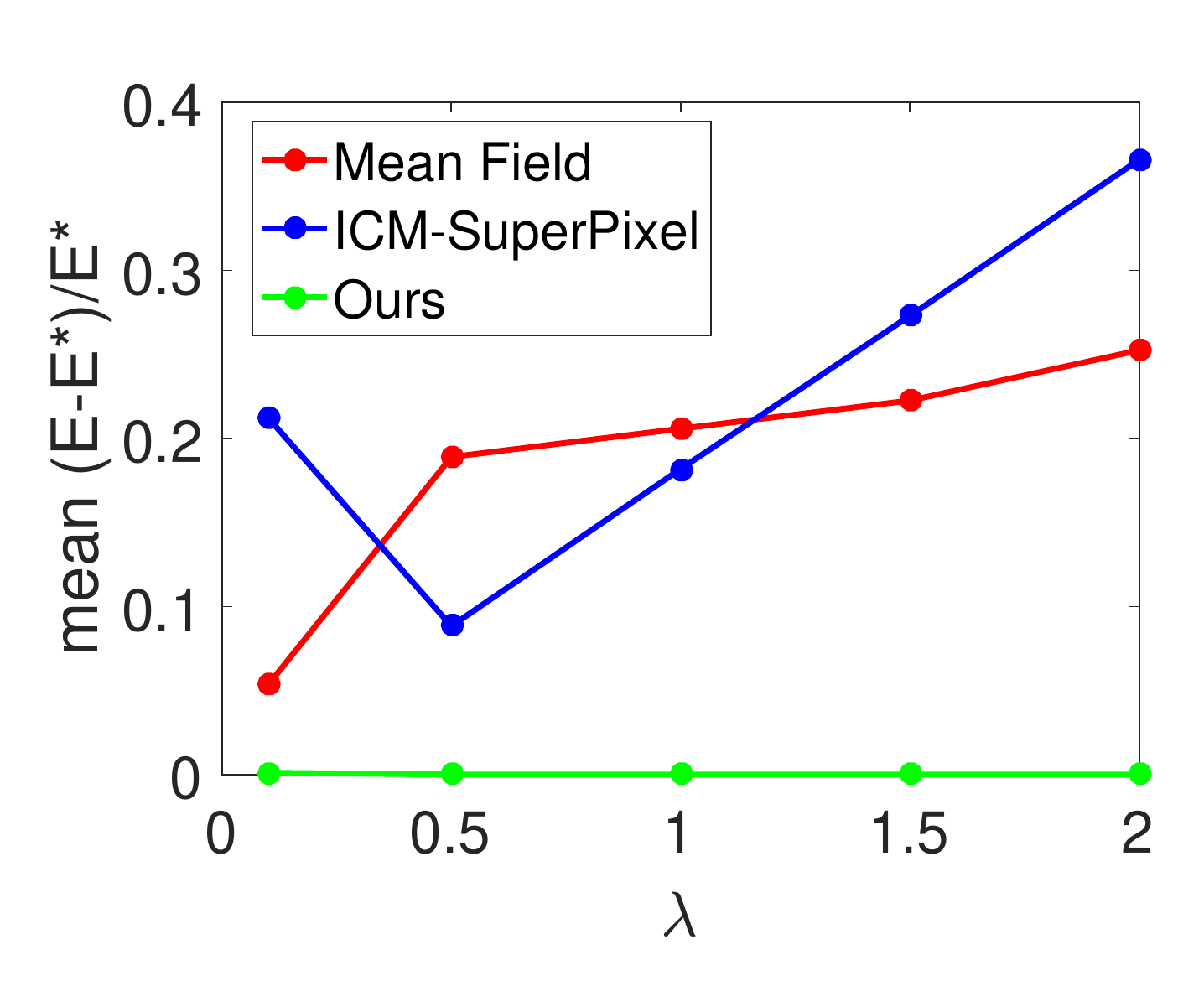}
 \includegraphics[trim={3cm 8cm 4cm 8cm},clip,width =0.23\textwidth]{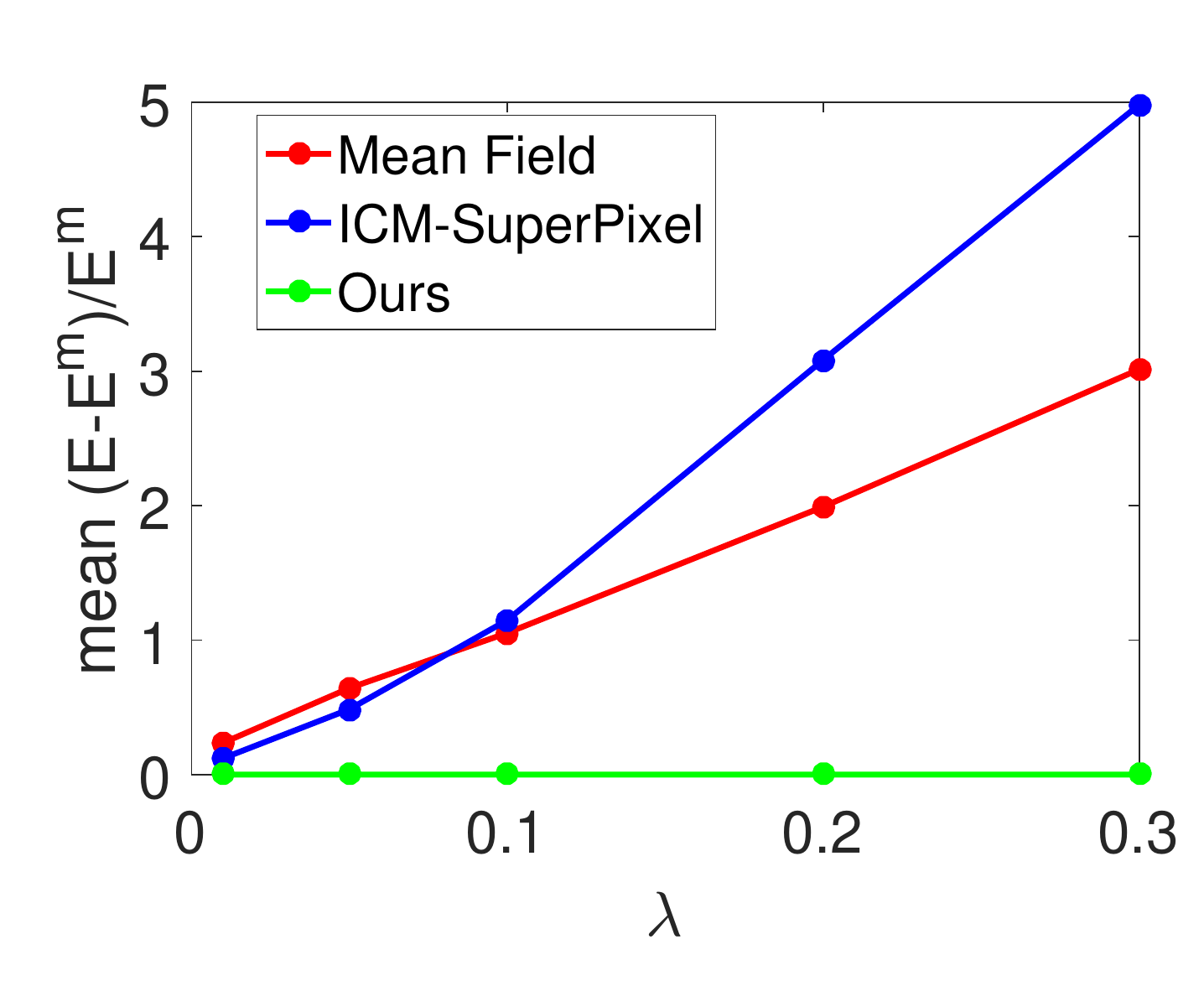}
\caption{Comparison of our method, superpixel ICM, and mean field. Left:  for binary Full-CRF; right: multilabel Full-CRF.  }
\label{fig:compEngBM}
\end{center}
\end{figure}

We now evaluate the binary optimization we develop in Sec.~\ref{sec:binary}. In this case, exact optimum can be computed
with a graph cut, but it is computationally prohibitive.
To make computation feasible, we
reduce the size of the images in PASCAL dataset to   $70$x$70$. To obtain a binary labeling problem, we choose the two most probable labels for each image. In particular, let $\bx^l$ be a labeling where each pixel is assigned label $l$. We find $l$ that gives the lowest value of
the energy, and $l'$ that gives the second lowest value.  Then for 
the binary energy in this section, $\calL = \{l,l'\}$. 
We compare our approach to exact optimization, mean-field, ICM, and superpixel-ICM. We also evaluated  jump moves instead of expansion moves for  optimizing the multi-label energy the original binary energy gets transformed to according to the approach in Sec.~\ref{sec:binary}.

We perform energy optimization with different settings of the smoothness parameter $\lambda$. Larger $\lambda$ 
correspond to energies that are more
difficult to optimize. In Fig.~\ref{fig:compEngBM}, left we show the results for mean field inference, superpixel ICM, and our method.  We omit the result of pixel ICM  and jump moves since they are significantly worse than the other methods.  

The optimal energy is computed with a graph cut. We plot the difference from the optimal energy normalized by the optimal energy, averaged over all images in the test dataset. In particular, if $E^*$ is the optimal
energy and $E$ is the energy returned by the algorithm, we compute $\frac{E-E^*}{E^*}$, the  energy increase relative to the optimal value. 
We average these relative increase values over  the validation fold of PASCAL VOC 2012 dataset.

Our method  finds the  globally optimal energy in the overwhelming majority of the cases, approximately 89\%. In the rest of the cases, the difference from the global optimum is tiny. The average relative energy increase 
is $ 0.00011$,  with a standard deviation of $  0.0012$. That is, on average, our algorithm
returns an energy worse than the optimal one by $0.011\%$. The maximum difference from the optimal energy value observed over the whole test dataset is $ 0.906$, and it happens   when the optimal energy value is $ 623.773$.  Thus for practical purposes we can say that our algorithm finds the global optimum for the binary Full CRF. 

Mean field inference works reasonably well for lower values of $\lambda$. For $\lambda = 0.1$, it finds an energy only about $5.3\%$ worse than the optimal one, on average.  Then the accuracy diminishes fast as $\lambda$ increases. For $\lambda = 2$, the average relative energy increase is close to $40\%$. 
Superpixel ICM is worse than mean field for smaller $\lambda$ values but outperforms mean field for larger values.  

The running times are  in Table~\ref{table:binaryTime}. The exact optimization is very expensive on these tiny images. 
Mean Field and Superpixel ICM are the most efficient, followed by  our approach.

\begin{table}
\begin{center}
\renewcommand\arraystretch{1.5}
\begin{tabular}{ |c|c| c|c|}
 \hline
   Mean Field  & Superpixel ICM & Ours & Exact\\
\hline
\hline
 $0.012 $& $0.014$ & $0.31 $& $7.1 $\\
 \hline
\end{tabular}
\vspace{1ex}
 \caption{Average running time  in seconds for the binary energy minimization.}\label{table:binaryTime}
\end{center}
\end{table}

\subsection{Multilabel Full CRFs}
We now compare our method with the mean field and superpixel ICM for the multi-label Full CRF energy. We omit pixel level
 ICM since it works significantly worse.
Again, we compare the energy optimization performance for different values of the smoothness parameter 
$\lambda$. In this case, the exact global optimum is not available. We still compare the relative
energy increase, but instead of the optimal energy, we use the smallest energy value found by any method. 
In all cases, our method has smaller than or equal energy than that of superpixel ICM and mean field.

Fig.~\ref{fig:compEngBM}, right, shows the relative energy increase plots for superpixel ICM, mean field, and our method.   Here $E^m$ 
stands for the smallest energy value obtained.   For small values of $\lambda$, all methods do well. For larger $\lambda$, superpixel ICM gets worse fast. This is because for larger $\lambda$ it tends to return the original labeling, unable to escape a bad local minimum. 
As $\lambda$ grows, the disparity in the performance increases even  more.
Thus the mean field inference is an appropriate inference method only if the unary terms are reliable, that is when there is no need to use a larger setting of $\lambda$. 
The running times are in Table~\ref{table:multiTime}.

\begin{figure*}
\begin{center}
 \includegraphics[width= 0.85\textwidth]{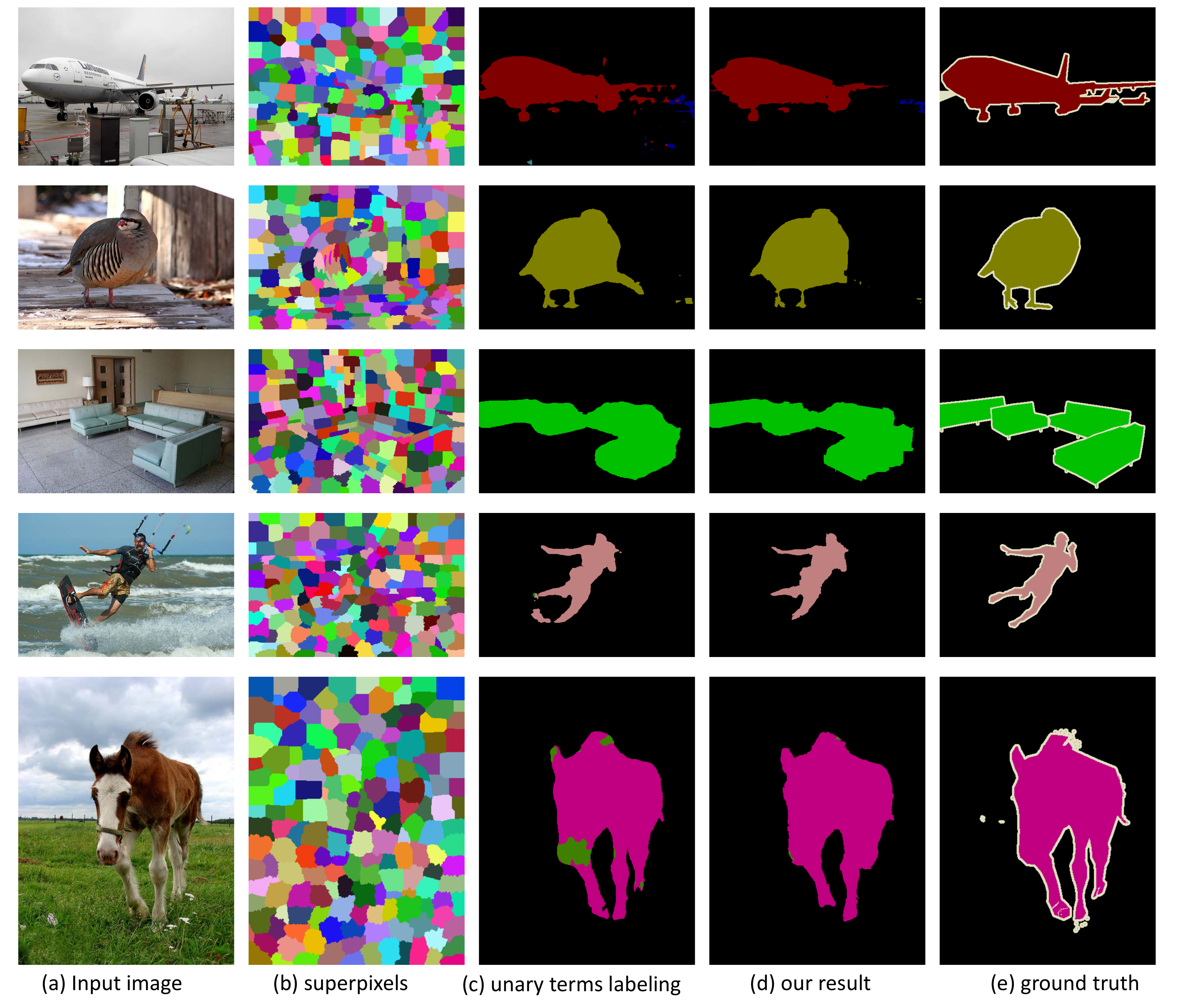}
\caption{A sample of results. 
 }
\label{fig:results}
\end{center}
\end{figure*}

\begin{table}
\begin{center}
\renewcommand\arraystretch{1.5}
\begin{tabular}{ |c|c| c|}
 \hline
   Mean Field  & Superpixel ICM & Our Method \\
\hline
\hline
 $2.16 $& $0.11 $ & $15.73$\\
 \hline
\end{tabular}
\vspace{1ex}
 \caption{Average running time in seconds for the multi-label energy minimization.}\label{table:multiTime}
\end{center}
\end{table}

\begin{table}
\begin{center}
\begin{tabular}{ |c|c|c|c|}
 \hline
object class &  Superpixels & Unary & Ours\\
 \hline
\hline
Overall &65.8899 &67.143&67.7484\\
 \hline
\hline
background&	91.996607 & 92.505& 	92.6236\\
aeroplane&	81.7341 &83.5563 &83.7498\\
bicycle	&41.1970 &51.1836 &51.2267\\
bird&	81.2498 &81.8296 &83.2405\\
boat&	58.5404 & 60.2947&60.1668\\
bottle&	58.4436 &59.62620&59.6262\\
bus	&79.8713 & 80.30270&81.0952  \\
car	&73.8574 & 75.22980&76.0474\\
cat	&78.1484 &	78.23960 &79.4247\\
chair&	26.9773 &27.49680&27.2861\\
cow	&65.7162 &66.69770 &67.5622\\
diningtable&	55.9211 &56.62960&56.6296\\
dog	 &68.5041 &69.3166 &69.9815\\
horse&	66.6537 & 66.9853&67.7631 \\
motorbike&	80.2764 &  	81.5684 &82.4261\\
person	&77.1641 &77.9252 &78.5284\\
pottedplant&	49.1919 & 49.65990&50.5761\\
sheep	&69.5786& 71.6253&71.9729\\
sofa	&42.1142&42.3743 &43.0364\\
train	&70.8761& 73.0517&73.4008\\
tvmonitor	&65.6757& 65.5707&66.3532\\
 \hline
\end{tabular}
\vspace{1ex}
 \caption{Results on PASCAL VOC 2012 Test data, using the  IOU measure.}\label{table:pascal}
\end{center}
\end{table}

\subsection{Semantic Segmentation Results}

Even though our primary goal is a more effective optimization algorithm, we also evaluate its usefulness for the task of semantic image segmentation.
Table~\ref{table:pascal} summarizes results on test PASCAL VOC 2012 set, using the Intersection over Union (IOU) measure. 
Using only unary terms (middle column) the IOU measure is $67.143$.  With our Full-CRF optimization (last column),
 the IOU measure goes up to $ 67.7484$. To insure that our improved results over the unary terms are not just due to superpixel tessellation, we also calculate the accuracy of the labeling based just on superpixels, without optimization. Namely we assign all pixels within the same superpixel the best single label that fits them. The IOU measure goes down to $65.8899$ (first column). For comparison, mean field optimization of our energy has a lower IOU measure of $67.3$.

A sample of results is shown in
figure~\ref{fig:results}. The average running time of our algorithm for these images was 15.73 seconds.

\section{Conclusion}
\label{sec:conclusion}
We introduced a new Full-CRF model with quantized edge weights, as well as an efficient  method for optimization in case of Potts  pairwise potentials. Our quantized edge model is
an approximation, and as such, offers insights into regularization properties of Gaussian edge Full CRF.  
In the case of binary Full-CRFs, our model experimentally produces a globally optimal solution in the overwhelming majority of the cases. In the multi-label case, we obtain significantly  better results compared to other frequently used methods, especially as the regularization strength is increased.  The main advantage of our model is that all edge weights are accounted for, no edge gets disregarded for the sake of approximation. We show the usefulness of our model and optimization for the task of semantic image segmentation.

\bibliographystyle{IEEEtran}
\bibliography{tpami17fullcrf}


\end{document}